\documentclass[dvipsnames, preprint, authoryear]{elsarticle}
\usepackage{graphicx}
\usepackage[english]{babel}
\usepackage[utf8]{inputenc}
\usepackage{booktabs}
\usepackage{courier}
\RequirePackage{amsmath,amsthm,amssymb,enumitem,graphicx,ifpdf}
\RequirePackage[left=1in,right=1in,top=1in,bottom=1in]{geometry}
\newcommand{\Z}{\mathbb{Z}}
\newcommand{\R}{\mathbb{R}}
\renewcommand{\Pr}{\mathbb{P}}

\usepackage{algorithm}
\usepackage{algpseudocode}
\usepackage{tikz}
\usetikzlibrary{shapes.geometric, shapes.misc, arrows}
\usepackage[colorlinks=true,linkcolor=black, citecolor=blue, urlcolor=blue]{hyperref}

\tikzstyle{start} = [rounded rectangle, minimum width=3cm, minimum height=1cm, text centered, draw=black, fill=RoyalBlue!10]
\tikzstyle{success} = [rounded rectangle, minimum width=2cm, minimum height=1cm,text centered, draw=black, fill=green!10, inner sep=5pt]
\tikzstyle{failure} = [rounded rectangle, minimum width=2cm, minimum height=1cm,text centered, draw=black, fill=red!10, inner sep=5pt]
\tikzstyle{decision} = [diamond, minimum width=3cm, minimum height=2cm, text centered, draw=black, aspect=1.3, inner sep=0pt]
\tikzstyle{process} = [rectangle, minimum width=3cm, minimum height=1cm, text centered, draw=black, inner sep=5pt]
\tikzstyle{arrow} = [thick,->,>=stealth]

\begin{document}
    
% \begin{titlepage}
%     \huge \centering Analyzing Customer-Facing Vendor Experiences with Time Series Forecasting and Monte Carlo Techniques
    
%     \vspace{\baselineskip}
%     \Large \centering
%     \begin{tabular}{c}
%         Jason Tang \\ \normalsize jastang@ebay.com 
%     \end{tabular}
%     %
%     \quad and \quad
%     %
%     \begin{tabular}{c}
%         Vivek Kaushik \\ \normalsize vikaushik@ebay.com 
%     \end{tabular}
    
%     \vspace{\baselineskip}
%     eBay Inc., 2025 Hamilton Ave., San Jose, CA 95125
% \end{titlepage}

\begin{frontmatter}
\title{Analyzing Customer-Facing Vendor Experiences with Time Series Forecasting and Monte Carlo Techniques \tnoteref{t1}}
% \renewcommand{\thefootnote}{\fnsymbol{footnote}}
% \author[1]{Jason Tang\thanks{Corresponding author} }
\author[eBay]{Vivek Kaushik}
\ead{vikaushik@ebay.com}
\author[eBay]{Jason Tang \corref{cor}}
\ead{jastang@ebay.com}
% \affil[1]{eBay Inc., 2025 Hamilton Ave., San Jose, CA 95125}
\address[eBay]{{eBay Inc., 2025 Hamilton Ave., San Jose, CA 95125}}
\cortext[cor]{Corresponding author}
\tnotetext[t1] {This research did not receive any specific grant from funding agencies in the public, commercial, or not-for-profit sectors}
\date{}

% \maketitle
% \renewcommand{\thefootnote}{\arabic{footnote}}

\begin{abstract}
eBay partners with external vendors, which allows customers to freely select a vendor to complete their eBay experiences. However, vendor outages can hinder customer experiences. Consequently, eBay can disable a problematic vendor to prevent customer loss. Disabling the vendor too late risks losing customers willing to switch to other vendors, while disabling it too early risks losing those unwilling to switch. In this paper, we propose a data-driven solution to answer whether eBay should disable a problematic vendor and when to disable it. Our solution involves forecasting customer behavior. First, we use a multiplicative seasonality model to represent behavior if all vendors are fully functioning. Next, we use a Monte Carlo simulation to represent behavior if the problematic vendor remains enabled. Finally, we use a linear model to represent behavior if the vendor is disabled. By comparing these forecasts, we determine the optimal time for eBay to disable the problematic vendor.
\end{abstract}


\end{frontmatter}

%%%%%%%%%%%%%%%%%%%%%%%%%%%%%%%%%%%%%%%%%%%%%%%

\section{Introduction}

In today’s landscape of rapid innovation, applications often use the services of third-party vendors to enable additional functionality quickly and easily. eBay is no exception. As an eCommerce company, eBay partners with many external vendors to serve various business functions. In some cases, multiple vendors support the same business function in order to provide the richest experience for all customers. 

As with any complex system, however, vendors are subject to outages during which they cannot operate normally. These outages could impact customers' ability to properly complete their experiences on eBay. Ideally, if a specific vendor is experiencing minimal impact during an outage, a customer using that vendor could have an initial failed experience with it, but should eventually have a successful experience upon retrying. 

\begin{figure}
    \centering
    \includegraphics[width=0.8\textwidth]{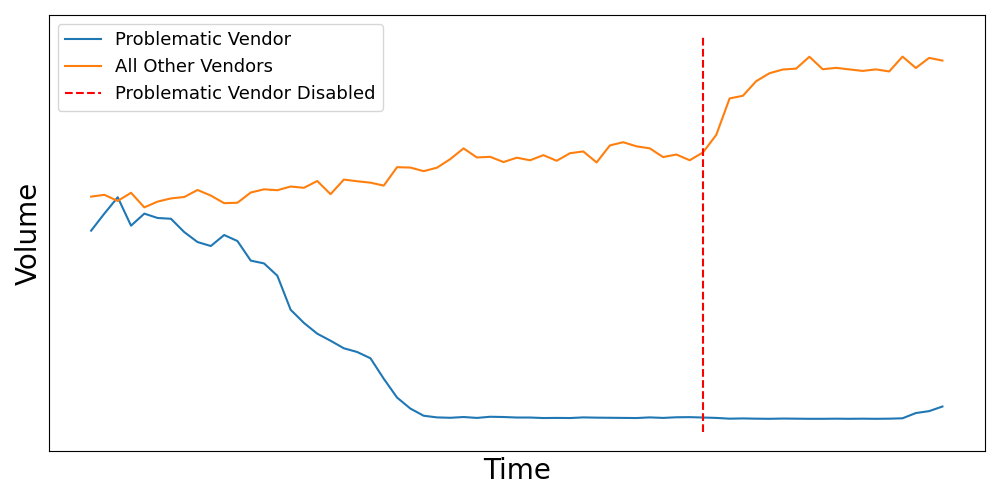}
    \caption{Real phenomena seen during an incident in which a problematic vendor was disabled. There are several features to note: (1) as the volume of experiences completed with the problematic vendor (blue) decreases, those completed with other vendors (orange) increases; (2) the increase in experiences completed with other vendors is delayed in time with respect to the decrease in volume for the problematic vendor  – it begins after the problematic vendor's volume starts decreasing and continues after the problematic vendor's volume stagnates; (3) we see an increase in experiences completed with other vendors after the problematic vendor is disabled (red dashed line). There are natural customer decisions corresponding to these phenomena: (1) as the problematic vendor's outage worsens, customers tend to migrate and complete their experiences with another vendor; (2) customers may retry with the problematic vendor several times before deciding to switch to another vendor, and these retries take a finite amount of time; (3) once the problematic vendor is disabled, more customers are willing to use a different vendor. We model all these effects in this paper.}
    \label{fig:example_incident}
\end{figure}

However, if the vendor is experiencing a high degree of impact, then customer retries may not be successful. Customers with unsuccessful retries may choose to switch to a different vendor to complete their experiences. If the vendor's impact is severe enough, then eBay can temporarily disable, or \textit{wire off}, the vendor, meaning customers would no longer see that vendor as an option. Consequently, more customers will choose to complete their experiences with a different vendor. At the same time, historical data has shown many customers have a very strong preference for using a particular vendor and will choose not to complete their experience if it is not shown as an option. We capture these effects in \textbf{Fig. \ref{fig:example_incident}}.

eBay's operations team makes the decision to wire off a problematic vendor. During past incidents, however, these decisions have not necessarily been data-driven. The decision to wire off a vendor is a critical one for the business; when it is executed, it typically is when the problematic vendor is experiencing a full outage, which means no customer requests with that vendor can successfully be served. During this type of situation, customers who prefer to only use the problematic vendor will not complete their experiences at all. On the other hand, customers who could otherwise use a different vendor are stuck retrying with the problematic vendor as long as it is being displayed to them. Over time, this results in net fewer successful customer experiences than if eBay had wired off the problematic vendor earlier during the incident.

In this paper, we present a data-driven approach to determine whether a given problematic vendor should be disabled and provide an optimal time for the wire-off. This optimal disabling time maximizes the total number of successfully completed customer experiences across all vendors. 

First, as an intermediary step, we forecast each vendor's \textit{baseline customer behavior}. More specifically, we forecast the total number of successful customer experiences for each vendor assuming all vendors are fully functioning. Historical data has shown baseline customer experience exhibits weekly seasonal behavior. Moreover, the seasonality varies with a trend. Therefore, for each vendor, we use a multiplicative seasonality model to represent its baseline customer behavior. We demonstrate how to forecast the seasonal and trend components associated with the baseline model, which in turn allows us to forecast each vendor's baseline customer behavior.

Next, we forecast the \textit{wired-on customer behavior}. More specifically, we forecast the total number of successful customer experiences across all vendors assuming the given problematic vendor remains enabled. In order to accomplish this, we first forecast the problematic vendor's \textit{availability}, which we interpret as the probability that a customer would have an initial successful experience with the problematic vendor. Historical incidents have shown that the problematic vendor's availability exhibits a decreasing trend over time; therefore, we use a Double Exponential Smoothing model to represent its availability. We describe the procedure to compute the smoothing and trend components associated with this model from historical data, which in turn allows us to forecast the problematic vendor's availability. Then, we consider several probability distributions that are constructed from historical incident data during which the problematic vendor remained enabled. These distributions model the following customer decisions: retrying the experience after an initial unsuccessful attempt with the problematic vendor, switching to another vendor, and the time between retries. We incorporate the baseline customer experience forecasts from the previous paragraph, the problematic vendor's availability forecast, and the aforementioned customer decision distributions into a Monte Carlo simulation. We run the Monte Carlo simulation to project future decisions made by customers pertaining to their experiences. Finally, we aggregate these simulation results together to forecast the wired-on customer behavior.

Lastly, we forecast the \textit{wired-off customer behavior}. More specifically, we forecast the total number of successful customer experiences across all enabled vendors assuming the given problematic vendor is disabled. Historical incidents have shown that customers who migrate from using the problematic vendor to using any other vendor typically do so at a constant rate over time. Therefore, we use a simple linear model to represent the wired-off customer experience. The independent variable in the model is the disabled vendor's predicted baseline customer experiences as described above, while the dependent variable is the difference between the total number of successful customer experiences during the wire-off and total number of predicted baseline customer experiences across all the enabled vendors. Forecasting the wired-off customer behavior amounts to determining the slope of the regression line.

Finally, in order to arrive at a data-driven decision for wiring off the given problematic vendor, we compare the wired-on and wired-off customer behavior forecasts from the previous two paragraphs. In particular, we check whether there is a future time at which the total number of predicted wired-off customer experiences begins to exceed the total number of predicted wired-on customer experiences. If such a time exists, then we choose to disable the problematic vendor at that particular time. Otherwise, we keep the problematic vendor enabled.

%%%%%%%%%%%%%%%%%%%%%%%%%%%%%%%%%%%%%%%%%%%%%%%

\section{Customer Behavior Models} \label{method}

Customer behavior is the underlying factor driving eBay's business metrics. In order to determine the optimal wire-off time for a given problematic vendor, we will need to consider models that represent distinct types of customer behavior. We use these models to forecast each vendor's \textit{volume}, which we interpret as the total number of customer experiences with that vendor. Throughout this section, we assume there are $N$ total vendors for some fixed positive integer $N.$

\subsection{Baseline Model} \label{baseline}

Consider the baseline situation in which all $N$ vendors are fully functioning, so that any customer is guaranteed to have an initial successful experience with their choice of vendor. We discuss how to forecast each vendor's baseline volume. For each $1 \le n \le N$ and each $t \ge 0$, we let $C_{n,t}$ be the $n$-th vendor's baseline volume at the time $t.$ Vendor volume is typically observed every minute, so we assume all times have granularity in minutes. In order to clearly distinguish between past, present, and future times, we will index time in the following way: we fix a $t_0$ to be the current time, and for each $m \in \Z,$ we let $t_m$ be the time corresponding to $m$ minutes since $t_0$. For any $m<0,$ we interpret $t_m$ as a historical time and for any $m>0,$ we interpret $t_m$ as a future time. We assume that volume for current and historical times is known, while volume for future times is unknown. We also assume that volume is positive for all times. 

\begin{figure}
    \centering
    \includegraphics[width=0.8\linewidth]{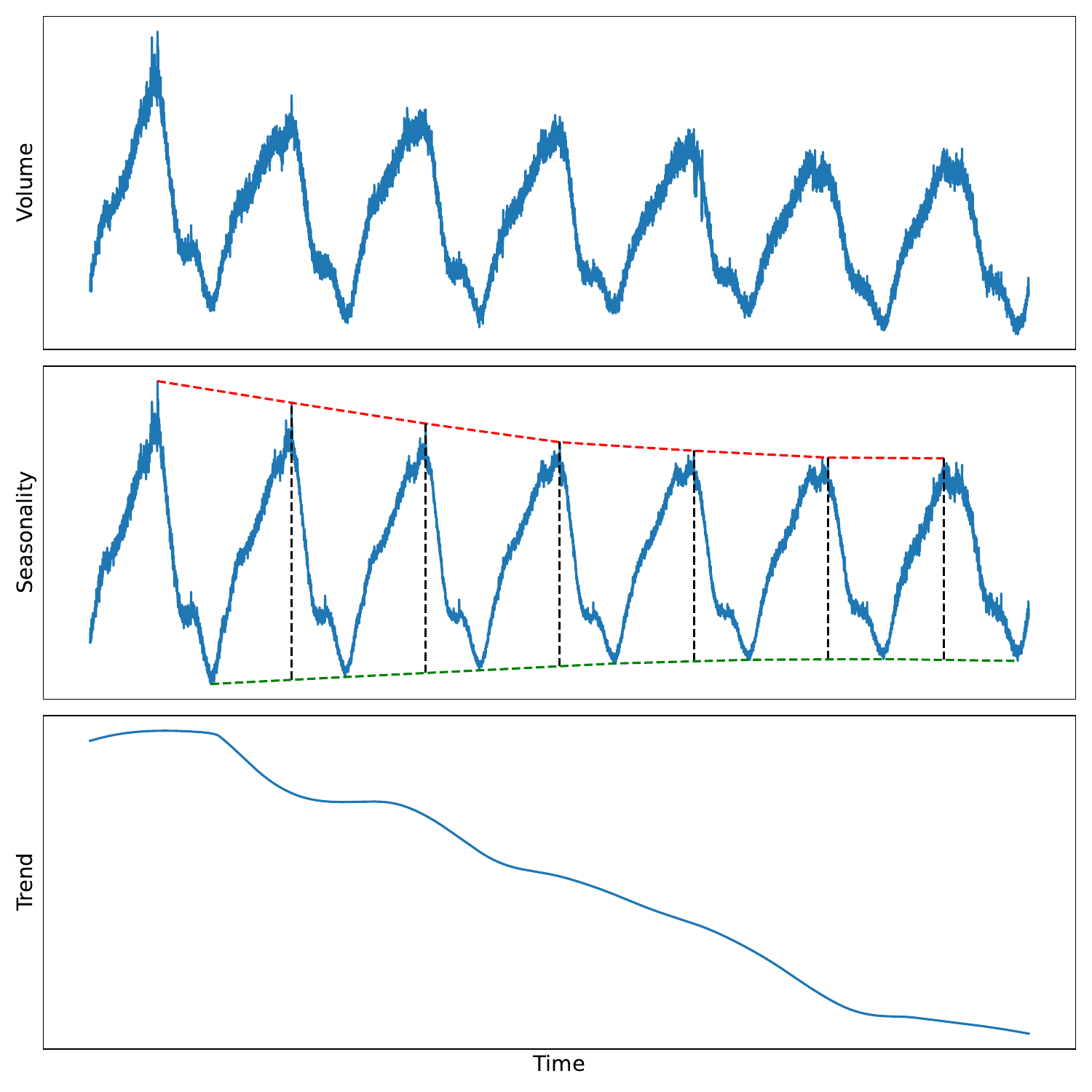}
    \caption{A realistic situation that strongly suggests the use of a multiplicative seasonality model. We plot the volume over time of a vendor (\textit{Top}). We perform an additive seasonal decomposition on the volume time-series to extract the individual seasonal component time-series (\textit{Middle}) and trend component time-series (\textit{Bottom}). We connect the approximate local maxima of the seasonal component time-series with a red dashed line and the approximate local minima with a green dashed line. We indicate the approximate seasonal fluctuation amplitudes with dashed black lines. We see these amplitudes and trend decrease throughout time. Therefore, we use a multiplicative seasonality model to represent this situation, as opposed to an additive seasonality model.}
    \label{fig:multiplicative_seasonality}
\end{figure}

Historical evidence has shown each vendor's baseline volume exhibits weekly seasonality along with trend. Moreover, due to varying customer activity levels throughout time, the seasonality directly varies with the trend (see \textbf{Fig. \ref{fig:multiplicative_seasonality}}). Therefore, we use a \textit{multiplicative seasonality model} \citep{multiplicativeSeasonality} to represent each vendor's baseline volume. Equivalently, we use an additive seasonality model to represent the \textit{logarithm} of each vendor's baseline volume. Explicitly, for the $n$-th vendor, we consider the model: 
    \begin{equation}\label{eq:Multiplicative Seasonality Model}
        \ln(C_{n,t_m}) = \ln( s_{n,t_m} ) + \ln(g_{n,t_m}) +\epsilon_{n,t_m},
    \end{equation}
where $s_{n,t_m}$ and $g_{n,t_m}$ are the seasonal and trend components at time $t_m$, respectively, and $\left \lbrace \epsilon_{n,t_m} \right \rbrace_{m \in \Z}$ is an independent and identically distributed sequence of Gaussian random variables with mean $0.$

\subsubsection{Seasonal Component}
We discuss how to forecast the $n$-th vendor's seasonal component in \textbf{Eq. \eqref{eq:Multiplicative Seasonality Model}}. To do this, we consider the following model for the logarithmic seasonal component:
    \begin{equation}\label{eq: Fourier Series}
        \ln(s_{n,t_m}) = \sum_{h=1}^{H_n} \alpha_{n,h} \cos \left({\frac{2 \pi   h t_m}{L}} \right) + \beta_{n,h} \sin \left({\frac{2 \pi    h t_m}{L}} \right) + \zeta_{n,t_m}
    \end{equation}
for each $m \in \Z,$ in which 
    \begin{itemize}
        \item $H_n$ is the known number of harmonics;
        \item $L=10080$ is the total number of minutes in a week;
        \item $\alpha_{n,h}$ and $\beta_{n,h}$ are the unknown Fourier coefficients for each $1 \le h \le H_n$;
        \item $\left \lbrace \zeta_{n,t_m} \right \rbrace_{m \in \Z}$ is an independent sequence of Gaussian random variables with mean $0$ and common variance $\sigma_n^2$ for some constant $\sigma_n>0.$
    \end{itemize}
More compactly, for each $m \in \Z,$ we can rewrite \textbf{Eq. \eqref{eq: Fourier Series}} as 
    \begin{equation} \nonumber
        \ln(s_{n,t_m}) = \mathbf x_{n,t_m}^\top \  \boldsymbol \beta_n  + \zeta_{n,t_m}, 
    \end{equation}
where
    \begin{align}
        \mathbf x_{n,t_m} &= \left( \cos \left(\frac{2 \pi (1) t_m}{L} \right) \ , \  \sin \left(\frac{2 \pi (1) t_m}{L} \right) \ , \ \dots \ , \ \cos \left(\frac{2 \pi (H_n) t_m}{L} \right) \ , \ \sin \left(\frac{2 \pi (H_n) t_m}{L} \right)  \right) \nonumber 
    \end{align}
is the vector of seasonality features at time $t_m$ and 
    \begin{align}
        \boldsymbol \beta_n &= (\alpha_{n,1} \ , \ \beta_{n,1} \ , \ \dots \ , \ \alpha_{n,H_n} \ , \ \beta_{n,H_n}) \nonumber.
    \end{align}
is the vector of unknown Fourier Coefficients.

Suppose $\mathbf s_{n,M} =\left \lbrace s_{n,t_m} \right \rbrace_{m = -M}^0 \in \R^{M+1}$ is a vector of seasonal component historical observations. Let $\ln(\mathbf s_{n,M}) = \left \lbrace \ln(s_{n,t_m}) \right \rbrace_{m = -M}^0$ be the vector as a result of taking the logarithm of each historical observation. Forecasting the seasonal component $s_{n,t_m}$ for $m>0$ amounts to determining the Fourier coefficient vector $\boldsymbol \beta_n.$ We utilize \textit{Maximum a Posteriori (MAP)} estimation, the main technique used in \citep{TaylorLetham2018}. The goal of MAP estimation is to seek the parameter $\boldsymbol \beta_n \in \R^{2H_n}$ maximizing $f \left(  \boldsymbol \beta_n \ | \ \ln(\mathbf s_{n,M}) \right),$ the conditional probability density function of the parameter $\boldsymbol \beta_n$ given the observed data. Equivalently, via Bayes Rule, we seek the parameter $\boldsymbol {\beta}_n$ maximizing the function $\phi: \R^{2H_n} \rightarrow \R$ defined by
    \begin{align} \label{eq: MAP Estimation}
        \phi(\boldsymbol \beta_n) &= f \left(   \ln(\mathbf s_{n,M})  \ | \ \boldsymbol \beta_n \right) \ f(\boldsymbol \beta_n),
    \end{align}
where $f(\boldsymbol \beta_n)$ is a specified prior probability density function of the parameter $\boldsymbol \beta_n.$ We assume $\boldsymbol \beta_n$ has a Gaussian prior distribution with mean $\mathbf 0$ and covariance matrix $\left(\sigma'_n \right)^2 \ \mathbf I_{2 H_n}$ for some fixed \textit{seasonality prior scale} $\sigma'_n>0$, where $\mathbf I_{2H_n}$ is the $2H_n \times 2H_n$ identity matrix. Moreover, by our model assumptions, $\ln(\mathbf s_{n,M})$ is Gaussian with mean $\mathbf X_{n,M} \  \boldsymbol \beta_n$ and covariance matrix $\sigma_n^2 \ \mathbf I_{M+1},$ in which $\mathbf X_{n,M}$ is the $(M+1) \times 2H_n$ matrix with $m$-th row equal to $\mathbf x_{n,t_{1-m}}^\top.$ Therefore, we can write out the function $\phi$ in \textbf{Eq. \eqref{eq: MAP Estimation}} explicitly:
    \begin{align}
        \phi(\boldsymbol \beta_n) &= \frac{\exp \left(-\frac{\| \ln\left(\mathbf s_{n,M} \right) - \mathbf X_{n,M} \  \boldsymbol \beta_n \|^2}{2 \sigma_n^2} \right)  \ \exp \left(- \frac{\| \boldsymbol \beta_n\|^2}{2 (\sigma_n')^2} \right)}{\left(2 \pi \right)^{\frac{M+2}{2}} \sigma_n \sigma'_n } \nonumber \\
        &= \frac{\exp \left(-\frac{\| \ln\left(\mathbf s_{n,M} \right) - \mathbf X_{n,M}\  \boldsymbol \beta_n \|^2}{2 \sigma_n^2} - \frac{\| \boldsymbol \beta_n\|^2}{2 (\sigma_n')^2} \right)}{\left(2 \pi \right)^{\frac{M+2}{2}} \sigma_n \sigma'_n } \nonumber.
    \end{align}
Computing the gradient of $\phi$ with respect to $\boldsymbol \beta_n,$ we obtain
    \begin{align}
         (\nabla \phi)(\boldsymbol \beta_n) &= -\left( \frac{  \mathbf X_{n,M}^\top \  \mathbf X_{n,M} \ \boldsymbol \beta_n - \mathbf X_{n,M}^\top \  \ln\left(\mathbf s_{n,M} \right) }{\sigma_n^2} + \frac{\boldsymbol \beta_n}{ \left(\sigma_n' \right)^2}\right) \ \phi(\boldsymbol \beta_n) \nonumber .
    \end{align}
Upon setting the gradient equal to $\mathbf 0$ and solving for $\boldsymbol \beta_n,$ we obtain the MAP estimate
    \begin{align}
        \widehat{\boldsymbol \beta}_n &= \left( \mathbf X_{n,M}^\top \ \mathbf X_{n,M} + \left(\frac{\sigma_n}{\sigma_n'} \right)^2 \mathbf I_{2H_n} \right)^{-1} \mathbf X_{n,M}^\top \ \ln \left(\mathbf s_{n,M} \right) \nonumber.
    \end{align}

Finally, for each $m>0,$ we compute $\widehat{s}_{n,t_m},$ the predicted seasonal component of the $n$-th vendor's baseline volume at time $t_m,$ by taking the inner product $\mathbf x_{n,t_m}^\top \  \widehat{\boldsymbol \beta}_n$ and exponentiating the result.

\subsubsection{Trend Component}

Next, we forecast the $n$-th vendor's trend component in \textbf{Eq.\ \eqref{eq:Multiplicative Seasonality Model}}. Let $\left \lbrace u_{n,d} \right \rbrace_{d=1}^{D_n}$ is a collection of $D_n$ known times during which the vendor's $n$-th trend growth rate changes. We refer to such times as \textit{trend changepoints}. Consequently, for each $m \in \Z,$ define $\mathbf a_{n,t_m} \in \R^{D_n}$ to be the vector whose $d$-th component is $1$ if $t_m \ge u_{n,d}$ and $0$ otherwise. We assume the trend component obeys a piecewise linear model of the form:
    \begin{equation}\label{eq:Piecewise Linear Trend}
        \ln(g_{n,t_m}) = \left(\kappa_n+\mathbf a_{n,t_m}^\top \ \boldsymbol \delta_n \right) t_m + \mathbf a_{n,t_m}^\top \boldsymbol \gamma_n +\theta_n + \zeta'_{n,t_m}
    \end{equation}
for each $m \in \Z,$ where 
    \begin{itemize}
        \item $\kappa_n \in \R$ is an unknown base growth rate;
        \item $\boldsymbol \delta_n \in \R^{D_n}$ is an unknown vector of growth rate adjustments;
        \item $\boldsymbol \gamma_n \in \R^{D_n}$ is an unknown vector of continuity correction factors;
        \item $\theta_n \in \R$ is an unknown base offset;
        \item $\left \lbrace \zeta'_{n,t_m} \right \rbrace_{m \in \Z}$ is an independent and identically distributed sequence of Gaussian random variables with mean $0.$
    \end{itemize}
In particular, the expression $\kappa_n+\mathbf a_{n,t_m}^\top \ \boldsymbol \delta_n$ appearing in \textbf{Eq. \eqref{eq:Piecewise Linear Trend}} is the sum of the base growth rate $\kappa_n$ and all $D_n$ growth rate adjustments at time $t_m.$ Meanwhile, the vector $\boldsymbol \gamma_n$ ensures the map $t \mapsto \ln(g_{n,t})$ is a continuous function and is determined once $\boldsymbol \delta_n$ is determined. 

Assume $\mathbf g_{n,M} = \left \lbrace g_{n,t_m} \right \rbrace_{m=-M}^0 \in \R^{M+1}$ is a vector of historical trend component observations, where $M$ is a fixed positive integer. Let $\ln(\mathbf g_{n,M}) = \left \lbrace \ln(g_{n,t_m}) \right \rbrace_{m=-M}^0$ be the vector formed by taking the logarithm of each historical trend component observation. In addition, let $\left \lbrace u_{n,d} \right \rbrace_{d=1}^{D_n}$ be a collection of known historical trend changepoints with $D_n<M+1$. Forecasting the trend component $g_{n,t_m}$ for each $m>0$ amounts to determining the parameters $\kappa_n, \boldsymbol \delta_n,$ and $\theta_n.$ We consider the MAP estimation technique, in which we seek the parameters maximizing the conditional probability density function $f\left(\kappa_n, \boldsymbol \delta_n, \theta_n \ |  \ \ln(\mathbf g_{n,M}) \right).$ Following \citep{TaylorLetham2018}, we impose the prior assumptions that the parameters $\kappa_n$ and $\theta_n$ are both Gaussian with mean $0,$ and \textit{changepoint prior scale} $\lambda_n>0,$ we impose the prior assumption $\boldsymbol \delta_n$ has Laplace distribution with mean $\mathbf{0}$ and covariance matrix $\lambda_n \ \mathbf I_{D_n}.$ We let $\widehat{\kappa}_n, \widehat{\boldsymbol \delta}_n, \widehat{\theta}_n$ be the resulting MAP estimates, and we set $\widehat{\boldsymbol \gamma}_n$ to have $d$-th component equal to $-u_{n,d} \  \delta_{n,d},$ for each $1 \le d \le D_n,$ where $\widehat{\delta}_{n,d}$ is $d$-th component of the MAP estimate $ \widehat{\boldsymbol \delta}_n.$

Now, for each $m>0,$ we compute $\widehat{g}_{n,t_m},$ the predicted trend component of the $n$-th vendor's baseline volume at future time $t_m.$ We randomly declare the time $t_m$ to be a trend changepoint with probability $\frac{D_n}{M+1}$ and not a trend changepoint with probability $1-\frac{D_n}{M+1}.$ For each $m>0,$ we randomly choose the future growth rate $\widehat{\delta}_{n,D_n+m}$ at time $t_m$ according to the aforementioned prior Laplace distribution if $t_m$ is labeled a trend changepoint and set $\widehat{\delta}_{n,D_n+m}=0$ otherwise. Consequently, we set $\widehat{\gamma}_{n, D_{n} +m }= - t_m \   \widehat{\delta}_{n,D_n+m}$ and compute
    \begin{align} 
        \ln(\widehat{g}_{n,t_m}) &= \left( \widehat{\kappa}_n + (\underbrace{1 \ , \ \dots \ , \ 1}_{\text{$D_n+m$ times}})^\top \ \left(\widehat{\boldsymbol \delta}_n    \ , \  \widehat{\delta}_{n,D_{n}+1} \ , \ \dots \ , \  \widehat{\delta}_{n,D_{n}+m} \right) \right) \ t_m  \nonumber \\
         & \qquad + \left( (\underbrace{1 \ , \ \dots \ , \ 1}_{\text{$D_n+m$ times}})^\top \ 
        \left(\widehat{\boldsymbol \gamma}_n  \ , \  \widehat{\gamma}_{n,D_{n}+1} \ , \ \dots \ , \  \widehat{\gamma}_{n,D_{n}+m} \right)  + \widehat{\theta}_n
        \right), \label{eq: Baseline Trend Component Forecast}
    \end{align}
where
    \begin{align*}
        \left(\widehat{\boldsymbol \delta}_n  \ , \  \widehat{\delta}_{n,D_{n}+1} \ , \ \dots \ , \  \widehat{\delta}_{n,D_{n}+m} \right)
    \end{align*}
is the $(D_n+m)$-dimensional vector formed by concatenating the MAP estimate $\widehat{\boldsymbol \delta}_n$ from the previous paragraph with the vector $\left(\widehat{\delta}_{n,D_{n}+1} \ , \ \dots \ , \  \widehat{\delta}_{n,D_{n}+m} \right),$ and $ \left(\widehat{\boldsymbol \gamma}_n  \ , \  \widehat{\gamma}_{n,D_{n}+1} \ , \ \dots \ , \  \widehat{\gamma}_{n,D_{n}+m} \right)$ is defined similarly. We take the exponential of \textbf{Eq. \eqref{eq: Baseline Trend Component Forecast}} to obtain the predicted trend component at time $t_m.$

Finally, for all $m>0,$ we obtain $\widehat{C}_{n,t_m},$ the  $n$-th vendor's predicted baseline volume at time $t_m,$ by multiplying the corresponding seasonal and trend component forecasts at time $t_m$ together:
    \begin{equation}\label{eq:Expected Baseline Forecast}
        \widehat{C}_{n,t_m}=\widehat {s}_{n,t_m} \ \widehat {g}_{n,t_m}.
    \end{equation} 
Henceforth, we refer to the computed number $\widehat{C}_{n,t_m}$ as the \textit{expected baseline volume} for vendor $n$ at time $t_m.$

\subsection{Wired-On Model} \label{wired-on-sims}

Suppose one particular vendor is experiencing an outage, but it still remains enabled to customers. Throughout this section, we let $1 \le n_0 \le N$ be the unique index corresponding to the problematic vendor. We describe our approach to forecasting the total \textit{wired-on volume}, which is the total volume across all $N$ vendors. First, we forecast the \textit{availability} of vendor $n_0$, which we interpret as the probability that a customer would have a successful experience with it on their first attempt. Then, we incorporate the $n_0$-th vendor's availability forecast and the expected baseline volume forecasts described in the previous section into a Monte Carlo simulation. At each future time, we use the Monte Carlo simulation to project the ultimate decision of a customer initially using the problematic vendor along with the time they make the decision. Finally, we aggregate the Monte Carlo simulation results together to forecast the wired-on volume.

\subsubsection{Vendor Availability Model} \label{availability_forecast}

Let $a_{n_0,t}$ be the problematic vendor's availability at time $t.$ Vendor availability is not necessarily known at future times, and there is often limited data from the past. Furthermore, when a vendor is experiencing issues, its availability typically exhibits a decreasing trend. Therefore,  we consider a \textit{Double Exponential Smoothing model} \citep{brown1957exponential, holt2004forecasting, NistDoubleExponential} to represent the problematic vendor's availability, which is explicitly
    \begin{align}
        S_{n_0,t_m} &= \alpha \  (a_{n_0,t_m}) + (1-\alpha)  \ (S_{n_0,t_{m-1}} + b_{n_0,t_{m-1}}) \label{eq: Double Exponential Smoothing Equation 1}\\
        b_{n_0,t_m} &= \eta  \  (S_{n_0,t_m} - S_{n_0,t_{m-1}}) + (1-\eta)  \ b_{n_0,t_{m-1}} \label{eq: Double Exponential Smoothing Equation 2},
    \end{align}
for each $m \in \Z,$ where
    \begin{itemize}
        \item $\alpha \in [0,1]$ is an unknown smoothing factor;
        \item $\eta \in [0,1]$ is an unknown trend factor;
        \item $S_{n_0,t_m} \in \R$ is an unknown smoothing component;
        \item $b_{n_0,t_m} \in \R$ is an unknown trend component;
    \end{itemize}

Let $\left \lbrace  a_{n_0,t_m}  \right \rbrace_{m=-M}^0$ be a historical sequence of the problematic vendor's availability observations for some $M>0.$ Forecasting the problematic vendor's availability $a_{n_0,t_m}$ for all $m>0$ amounts to determining the factors $\alpha$ and $\eta$ based on the historical observations. First, we initialize the smoothing component in \textbf{Eq. \eqref{eq: Double Exponential Smoothing Equation 1}} and the trend component in \textbf{Eq. \eqref{eq: Double Exponential Smoothing Equation 2}} by setting
    \begin{align}
        S_{n_0,t_{-M}} &:=a_{n_0,t_{-M}} \nonumber \\
        b_{n_0,t_{-M}} &:=a_{n_0,t_{-M+1}}-a_{n_0,t_{-M}} \nonumber.
    \end{align}
With these initialized values, we seek the optimal factors $\alpha$ and $\eta$ that minimize the quantity
    \begin{align} \label{eq: MSE}
 \frac{\left \| \left \lbrace a_{n_0,t_m} - S_{n_0,t_m} -b_{n_0,t_m} \right \rbrace_{m=-M}^0  \right \|_2}{\sqrt{M+1}} &=  \sqrt{\frac{1}{M+1} \sum_{m=-M}^0 \left(a_{n_0,t_{m}} - S_{n_0,t_{m}} -b_{n_0,t_m} \right)^2},
    \end{align}
the \textit{root mean-squared error} between $\left \lbrace  a_{n_0,t_m}  \right \rbrace_{m=-M}^0$ and $\left \lbrace S_{n_0,t_{m}} +b_{n_0,t_m}  \right \rbrace_{m=-M}^0$. The root mean-squared error is a continuous function of the tuple $(\alpha,\eta)$ with domain equal to the unit square $[0,1]^2.$ By the extreme value theorem, we are guaranteed optimal factors $\alpha$ and $\eta$ that minimize the root mean-squared error. Seeking such factors, however, is a nonlinear optimization problem that cannot be solved analytically. Nevertheless, if we can determine these optimal factors, then we can compute $\widehat{a}_{n_0,t_m},$ the problematic vendor's predicted availability at time $t_m$ for each $m>0,$ via the formula
    \begin{equation} \label{eq:Availability Forecast}
        \widehat{a}_{n_0,t_m} = S_{n_0,t_{0}} +  m \  b_{n_0, t_{0}}.
    \end{equation}

\subsubsection{Customer Behavior Monte Carlo Simulation} \label{probabilistic_model}

We now describe the Monte Carlo simulation to forecast customer behavior when the problematic vendor remains enabled. Recall $n_0$ is the index corresponding to the problematic vendor. The Monte Carlo simulation combines together each vendor's expected baseline volume forecasts, the problematic vendor's availability forecast, and certain probability distributions modeling specific customer decisions. We compute these distributions over a historical time period during which the problematic vendor had previously experienced issues.

First, for each $k \ge 0,$ we consider the \textit{retry distribution}, which is a Bernoulli distribution with parameter
    \begin{equation} \label{eq:Retry Distribution}
        \pi_{n_0,k} = \Pr \left(\text{a customer retries with \textit{any} vendor} \ | \ \text{customer has failed $k$ times with $n_0$} \right),
    \end{equation}
where $\Pr$ stands for probability. The second distribution is the \textit{switch distribution}, which is another Bernoulli distribution with parameter \begin{equation}\label{eq:Switch Distribution}
        \rho_{n_0,k} = \Pr(\text{customer switches to another vendor} \ | \ \text{customer is retrying after failing $k$ times with $n_0$}).
    \end{equation}
Lastly, we consider the \textit{interattempt time distribution}, which has cumulative distribution function
    \begin{equation} \label{eq:Interattempt Time Distribution}
        \tau_{n_0}(s) = \Pr \left(\text{customer takes at least $s$ seconds to retry with any vendor} \ | \  \text{customer failed with $n_0$} \right),
    \end{equation}
where $s$ is a non-negative integer.

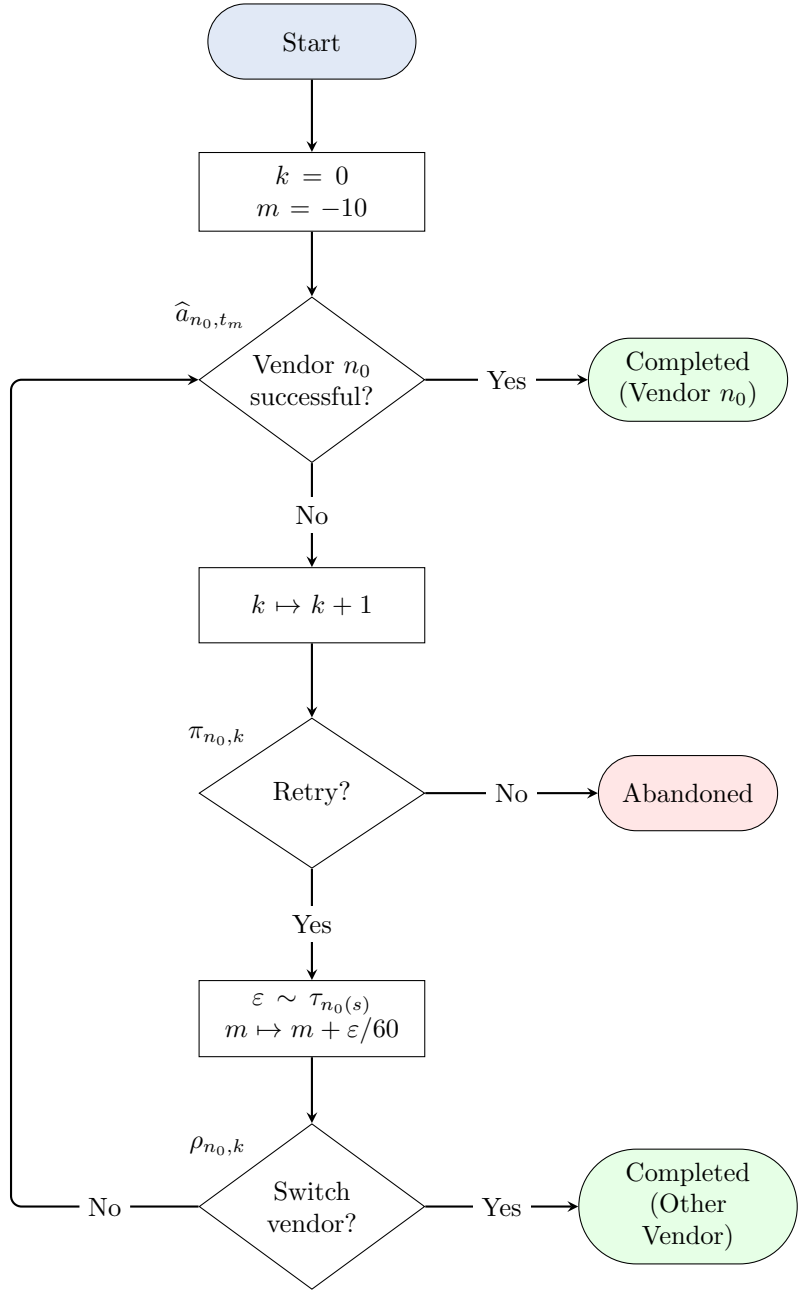
\begin{figure}
    \centering
    \begin{tikzpicture}[node distance=4cm]
        % Define nodes for the flow chart
        \node(start)[start]{Start};
        \node(initialize)[process, below of=start, text width=2cm, yshift=2cm]{$k=0$ \\ $m=-10$};
        \node(vendor0)[decision, below of=initialize, text width=2cm, yshift=1.5cm, label=north west:{$\widehat{a}_{n_0,t_m}$}]{Vendor $n_0$ successful?};
        \node(updatek)[process, below of=vendor0, text width=2cm, yshift=1cm]{$k \mapsto k + 1$};
        \node(retry)[decision, below of=updatek, text width=2cm, yshift=1.5cm, label=north west:{$\pi_{n_0, k}$}]{Retry?};
        \node(update)[process, below of=retry, text width=2.5cm, yshift=1cm]{ $\varepsilon  \sim \tau_{n_0(s)}$ \\ $m \mapsto m + \varepsilon/60$};
        \node(switch)[decision, below of=update, text width=2cm, label=north west:{$\rho_{n_0, k}$}, yshift=1.5cm]{Switch vendor?};
        \node(completed)[success, right of=vendor0, text width=2cm, xshift=1cm]{Completed (Vendor $n_0$)};
        \node(abandoned)[failure, right of=retry, text width=2cm, xshift=1cm]{Abandoned};
        \node(other)[success, right of=switch, text width=2cm, xshift=1cm]{Completed (Other Vendor)};
        
        % Dummy nodes for the wraparound arrow
        \node(ctrl1)[left of=switch]{};
        \node(ctrl2)[left of=vendor0]{};
        
        % Arrows for the flow chart
        \draw[arrow](start) -- (initialize);
        \draw[arrow](initialize) -- (vendor0);
        \draw[arrow](vendor0) -- node[fill=white]{Yes} (completed);
        \draw[arrow](vendor0) -- node[fill=white]{No} (updatek);`
        \draw[arrow](updatek) -- (retry);
        \draw[arrow](retry) -- node[fill=white]{No} (abandoned);
        \draw[arrow](switch) -- node[fill=white]{Yes} (other);
        \draw[arrow](retry) -- node[fill=white]{Yes} (update);
        \draw[arrow](update) -- (switch);
        \draw[arrow, rounded corners](switch) -- node[fill=white]{No} (ctrl1.center) -- (ctrl2.center) -- (vendor0);
    \end{tikzpicture}
    \caption{Decision flowchart showing the possible decisions a customer can make depending on whether they succeed with using the problematic vendor $n_0$ on their initial experience. Here, $m$ is initialized to $-10$. The Monte Carlo simulation, based on this flowchart, projects the ultimate decision (listed in a red or green box) a customer makes. The integer part of the final value of the index $m$ represents the time taken for the customer to make their decision.} 
    \label{fig:flowChart}
\end{figure}

Given a customer who initially attempts to use vendor $n_0$ at time $t_m$ for some integer $m,$ we describe the Monte Carlo simulation to project their ultimate decision, which we illustrate in \textbf{Fig. \ref{fig:flowChart}}. We initialize the the number of failures $k = 0$. In order to capture time delayed effects during our actual time of interest, we initialize $m$ to be some negative integer. We simulate the outcome (success or failure) of the customer's initial attempt with the problematic vendor according to the Bernoulli distribution with parameter $\widehat{a}_{n_0,t_m}$ as defined in \textbf{Eq. \eqref{eq:Availability Forecast}}. When $m \le 0$, we use the problematic vendor's actual availability $a_{n_0,t_m}$ in place of the predicted availability. If the customer succeeds, then we record that outcome at time $t_m$ and terminate the simulation. Otherwise, we increment $k$ and simulate the customer's decision whether to retry the experience with any vendor based on the retry distribution defined in \textbf{Eq. \eqref{eq:Retry Distribution}}. If the customer decides to abandon the experience, we record their decision and terminate the simulation. Otherwise, we simulate the customer's decision whether to switch to another vendor based on the switch distribution defined in \textbf{Eq. \eqref{eq:Switch Distribution}}. We also simulate the time $\varepsilon$ (in seconds) taken for the customer to execute their decision based on the aforementioned interattempt time distribution defined in \textbf{Eq. \eqref{eq:Interattempt Time Distribution}}. We terminate the simulation if the customer switches to another vendor and record this decision at time $t_{ \left \lfloor m+ 
\frac{\varepsilon}{60} \right \rfloor}$. Otherwise, we repeat the entire  procedure, but we replace $m$ with $m+
\frac{\varepsilon}{60}$. This process is described in \textbf{Algorithm \ref{alg:simulation}}.

Finally, we forecast the total wired-on volume by aggregating the Monte Carlo simulation results from the previous paragraph. Recall that $\widehat{C}_{n_0,t_m}$ is the problematic vendor's baseline volume at time $t_m,$ as defined in \textbf{Eq. \eqref{eq:Expected Baseline Forecast}}. Here, we interpret the integer $\left \lfloor \widehat{C}_{n_0,t_m} \right \rfloor$ as the predicted total number of customers who use the problematic vendor at time $t_m.$ For each $1 \le i \le \left \lfloor \widehat{C}_{n_0,t_m}\right \rfloor,$ we project the $i$-th customer's ultimate decision and their time of decision based on the Monte Carlo simulation. Now, we let $\widehat{A}_{n_0,t_m}$ be the total number of customers projected to have successful experiences with the problematic vendor at time $t_m$. We also let $\widehat{A}_{\text{other},t_m}$ be the total number of customers projected to have successful experiences with other vendors at time $t_m.$ Lastly, we let $\widehat{C}_{\text{other},t_m}$ be the total sum of all expected baseline volume forecasts across all vendors excluding the problematic vendor. Using these predicted numbers, we compute $\widehat{W}_{n_0,t_m,\text{on}},$ the predicted wired-on volume at time $t_m,$ via the formula
    \begin{equation}\label{eq:Wired-on Forecast}
        \widehat{W}_{n_0,t_m,\text{on}} = \widehat{A}_{n_0,t_m} + \widehat{A}_{\text{other},t_m} +\widehat{C}_{\text{other},t_m}.
    \end{equation}
%

% predicted Total Carts Wired on = Simulated Additional Vendor + Simulated Additional Other + expected baseline other

\begin{algorithm}
\caption{Customer Decision Simulation}
\label{alg:simulation}
\begin{algorithmic}[1]
\Require $\widehat{a}_{n_0,t}$ \Comment{Probability of initial success using vendor $n_0$ at time $t$}
\Require $\pi_{n_0,k}$ 
\Comment{Probability of retrying with any vendor given $k$ failures with vendor $n_0$}
\Require $\boldsymbol \tau_{n_0}$ 
\Comment{Interattempt time distribution probability vector having $s$-th component $\tau_{n_0}(s)$}
\Require $\rho_{n_0,k}$
\Comment{Probability of switching vendors on a retry after $k$ failures with vendor $n_0$}
\newline

\Ensure Final Status (success with $n_0$, success with another vendor, or complete abandonment) with time

\Procedure{customer decision}{$m$}
\State $k \gets 0$

\While{$k \le 15$}
    \State $u \gets \text{Uniform}(0,1).$ 
\If{$u \leq \widehat{a}_{n_0,t_{\lfloor m \rfloor}}$}
    \State \Return ``Success with vendor $n_0$," $t_{\lfloor m \rfloor}$

\Else 
    \State $k \gets k+1$
    \State $u \gets \text{Uniform}(0,1).$
    \If{$u \ge \pi_{n_0,k}$ or $k=15$}
        \State \Return ``Abandoned," $t_{\lfloor m \rfloor}$
    \Else
        \State $\varepsilon \gets \tau_{n_0}$ 
        \State $m \gets m + \frac{\varepsilon}{60}$
        \State $u \gets \text{Uniform}(0,1).$
        \If{$u \leq \rho_{n_0,k}$}
        \State \Return ``Success with another vendor," $t_{\lfloor m \rfloor}$
        \EndIf

    \EndIf
\EndIf
\EndWhile
\State \Return ``Abandoned," $t_{\lfloor m \rfloor}$
\EndProcedure%

\end{algorithmic}
\end{algorithm}

%https://www.overleaf.com/learn/latex/Algorithms

\subsection{Wired-Off Model} \label{wired-off-sims}
\begin{figure}
    \centering
    \includegraphics[width=0.49\textwidth]{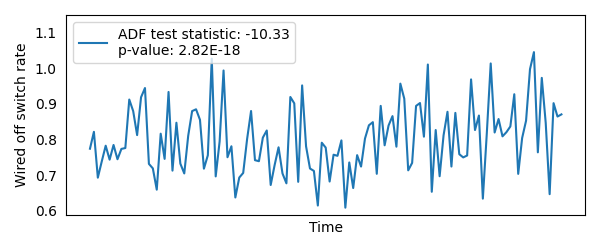}
    \includegraphics[width=0.49\textwidth]{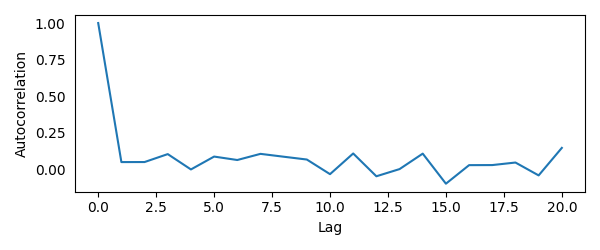}
    \caption{The proportion of the problematic vendor's baseline volume which contributed to the additional volume for all other enabled vendors (\textit{left}) and its corresponding auto-correlation function (\textit{right}). When performing an Augmented Dickey-Fuller test on this ratio over time, we get a test statistic of -10.33 which corresponds to a $p$-value of $2.82 \times 10^{-18}$, suggesting that this series is stationary. We can confirm this by plotting the auto-correlation function, which rapidly decays.}
    \label{fig:stationarity}
\end{figure}

Now, suppose the problematic vendor $n_0$ is disabled. Let $W_{n_0,t_m, \text{off}}$ be the total \textit{wired-off volume} at time $t_m$, which is the total volume across all enabled vendors at time $t_m$.  In this situation, customers will not be able to retry their experiences with the problematic vendor. Past incident data has shown that the ratio of additional volume for vendors other than the problematic one to the baseline volume of the problematic vendor is stationary over time (\textbf{Fig. \ref{fig:stationarity}}); we performed an Augmented Dickey-Fuller test \citep{Greene_2003} which resulted in a $p$-value of $2.82 \times 10^{-18}$. Therefore, we do not need to consider any time-dependent effects of the rate at which customers who typically use the problematic vendor would switch to use another vendor. We consider a simple linear model of the form
    \begin{equation}\label{eq:Wired Off Linear Model}
       W_{n_0,t_m, \text{off}}  = \Delta_{n_0} \widehat{C}_{n_0,t_m} + \widehat{C}_{\text{other},t_m} + \epsilon_{n_0, t_m},
    \end{equation}
to represent the wired-off volume, where for each $m \in \Z,$ 
    \begin{itemize}
        \item $\widehat{C}_{n_0,t_m}$ is the disabled vendor's expected baseline volume at time $t_m,$ as defined in \textbf{Eq. \eqref{eq:Expected Baseline Forecast}};

        \item $\widehat{C}_{\text{other},t_m}$ is the total expected baseline volume across all enabled vendors at time $t_m$;

        \item $\Delta_{n_0} \in \R$ is the unknown slope;

        \item $\left \lbrace \epsilon_{n_0,t_m}\right \rbrace_{m \in \Z}$ is an independent and identical sequence of Gaussian random variables with mean $0;$
    \end{itemize}

Forecasting the wired-off volume amounts to determining the slope $\Delta_{n_0}$ based on historical data over a timeframe during which problematic vendor $n_0$ was disabled. To this end, let $T_\text{hist}$ be a finite collection of historical times corresponding to a previous known incident during which the problematic vendor was disabled. Suppose we have a vector $\mathbf W_{n_0, T_\text{hist}, \text{off} } = \left \lbrace W_{n_0,t, \text{off}} \right \rbrace_{t \in T_\text{hist} }$ containing the total wired-off volume observations over $T_{\text{hist}}.$ Via \textbf{Eq. \eqref{eq:Expected Baseline Forecast}}, we compute the predicted vector $\widehat{\mathbf C}_{n_0,T_\text{hist} } = \left \lbrace \widehat{C}_{n_0,t} \right \rbrace_{t \in T_\text{hist} }$ containing the expected baseline volume observations corresponding to the disabled problematic vendor. Then, we compute the predicted vector $\widehat{\mathbf C}_{\text{other},T_\text{hist} } =  \left \lbrace \sum_{n \ne n_0} \widehat{C}_{n,t} \right \rbrace_{t \in T_\text{hist} },$ in which each observation at time $t$ is the sum of forecasted baseline volumes at time $t$ across all enabled vendors. Now, we can obtain an estimate for the slope $\Delta_{n_0}$ by solving the linear equation 
    \begin{equation}\label{eq:Wired Off Linear Model Vector Form}
        \mathbf{W}_{n_0, T_\text{hist}, \text{off} } = \Delta_{n_0}\widehat{\mathbf C}_{n_0,T_\text{hist} } + \widehat{\mathbf C}_{\text{other},T_\text{hist} },
    \end{equation}
for $\Delta_{n_0}.$ Upon subtracting  $\widehat{\mathbf C}_{\text{other},T_\text{hist} }$ from both sides of \textbf{Eq. \eqref{eq:Wired Off Linear Model Vector Form}}, left multiplying both sides by $\widehat{\mathbf C}_{n_0,T_\text{hist} }^\top$, and finally dividing both sides of the equation by $\left(\widehat{\mathbf C}_{n_0,T_\text{hist} } \right) ^ \top \widehat{\mathbf C}_{n_0,T_\text{hist} } = \|\widehat{\mathbf C}_{n_0,T_\text{hist} }\|_2^2$, we obtain the estimate
    \begin{align}
        \widehat{\Delta}_{n_0} 
        &= \frac{\left(\widehat{\mathbf C}_{n_0,T_\text{hist} } \right)^\top  \left(\mathbf{W}_{n_0, \text{off}, T_\text{hist}  } - \widehat{\mathbf C}_{\text{other},T_\text{hist} }\right) }{\|\widehat{\mathbf C}_{n_0,T_\text{hist} }\|_2^2}.
    \end{align}
Thus, for all $m>0,$ we can compute $\widehat{W}_{n_0,t_m, \text{off}},$ the predicted wired-off volume at time $t_m,$ via the formula
    \begin{equation}
        \widehat{W}_{n_0,t_m, \text{off}} = \widehat{\Delta}_{n_0} \widehat{C}_{n_0,t_m} + \widehat{C}_{\text{other},t_m}.
    \end{equation}

% During a historical incident timeframe in which vendor $n$ previously was disabled, we compute $\widehat{p}_{\text{hist}}$, the mean proportion  of customers who switched from $n$ to another vendor. So for each $-10 \leq m \leq 2,$ we compute 
% %
%     \begin{equation}\label{eq:Wired-off forecast}
%         \widehat{W}_{n,t_m, \text{off}} = \widehat{p}_{\text{hist}} \left(\widehat{C}_{n,t_m} \right) +\widehat{C}_{\text{other},t_m},   
%     \end{equation} 
% %
% representing the total predicted volume assuming $n$ is unavailable to customers.

%%%%%%%%%%%%%%%%%%%%%%%%%%%%%%%%%%%%%%%%%%%%%%%

\section{Results and Validation}

In this section, we discuss sample results and the validation of our forecasting models on vendor volume data. We apply our methodology to real incidents and analyze customer and business benefits.

\subsection{Expected Baseline Volume Forecast Validation}

For each $1 \le n \le N,$ recall that we represent the $n$-th vendor's expected baseline volume through the additive seasonal model for its logarithm as defined in 
\textbf{Eq. \eqref{eq:Multiplicative Seasonality Model}}. In order to forecast the baseline volume, we leverage \textit{Prophet}, a forecasting library developed by Meta. In \textbf{Section \ref{baseline}}, we discussed how to forecast the trend and seasonal components individually via MAP estimations given historical data associated with such components. For computational efficiency, however, Prophet computes a single MAP estimate for the concatenated list of parameters $\Phi_n=(\boldsymbol \beta_n, \kappa_n, \boldsymbol \delta_n, \theta_n),$ in which we recall $\boldsymbol \beta_n$ is the vector of Fourier Coefficients in the seasonal component model, and $(\kappa_n, \boldsymbol \delta_n, \theta_n)$ is the list of parameters in the trend component model. This single MAP estimate for $\Phi_n$ is computed based on the $n$-th vendor's historical baseline volume data.

\begin{figure}
    \centering
    \includegraphics[width=0.8\textwidth]{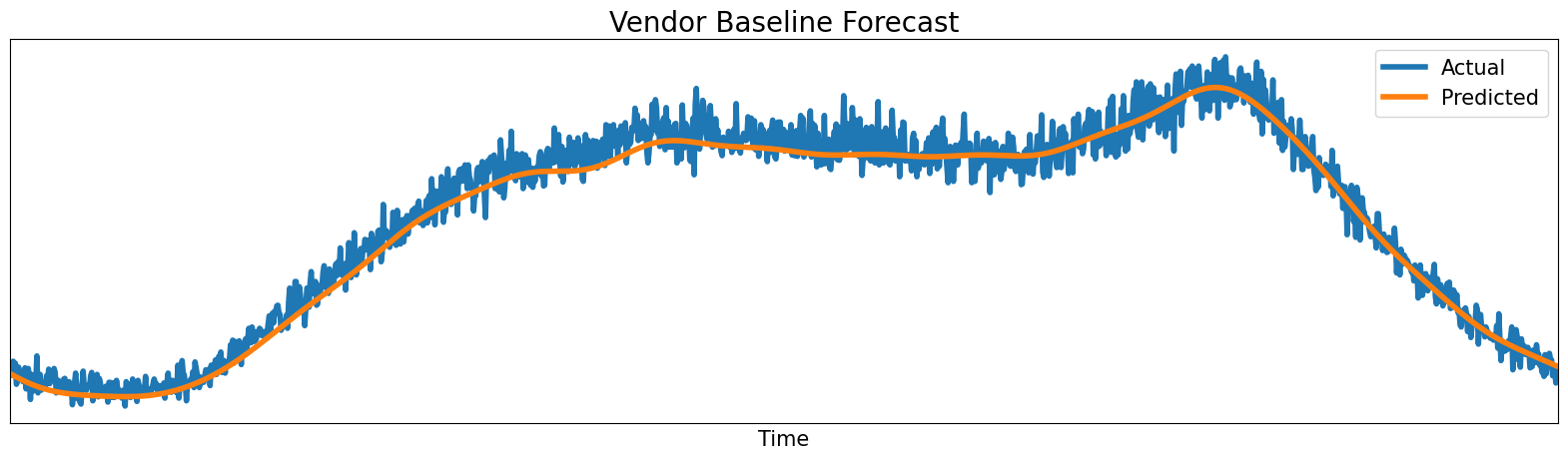}
    \caption{Forecasting the baseline volume for a sample vendor using the multiplicative weekly seasonality model. We plot the true baseline volume in blue and our predicted baseline volume in orange.}
    \label{fig:baseline_forecast}
\end{figure}

Moreover, Prophet exposes several hyperparameters to control model fitting. We list the main ones that affect forecasting in \textbf{Table \ref{tab:hyperparameters}}. The default values of such hyperparameters, however, typically do not result in a good fit for vendor data. Therefore, we employ an efficient random hyperparameter search \citep{BergstraBengio} to determine an optimal set of hyperparameters that minimizes the root mean-squared error between the actual future baseline volume $\left \lbrace C_{n,t_m} \right \rbrace_{m \ge 0}$ and the expected baseline volume $\left \lbrace \widehat{C}_{n,t_m} \right \rbrace_{m \ge 0}$ computed via \textbf{Eq. \eqref{eq:Expected Baseline Forecast}}. We list the hyperparameters that we tune, along with their lower and upper bounds over our search space, in  \textbf{Table \ref{tab:hyperparameters}}. Furthermore, in \textbf{Fig. \ref{fig:baseline_forecast}}, we display an example of forecasting a vendor's baseline volume based on two weeks of past historical volume data as a result of the hyperparameter grid search, computing the MAP estimate for the concatenated list of parameters $\Phi_n,$ and using \textbf{Eq. \eqref{eq:Expected Baseline Forecast}}.

\begin{table}
    \centering
    \begin{tabular}{|c c c|}
        \hline
        Hyperparameter & Lower bound & Upper bound  \\            
        \hline \hline
        \texttt{seasonality\_prior\_scale} & 0.01 & 10  \\
        \hline
        \texttt{changepoint\_prior\_scale} & 0.001 & 1 \\ 
        \hline 
        \texttt{weekly\_seasonality} & 10 & 30 \\
        \hline
    \end{tabular}
    \caption{The multiplicative seasonality model hyperparameters and their corresponding bounds we consider to forecast the expected baseline volume for each vendor. The seasonality and changepoint prior scale hyperparameters measure the influence of seasonality and changepoints on the model fit. Finally, the weekly seasonality hyperparameter describes how many harmonics are in the Fourier series having period equal to one week. Both prior scales are tuned on a logarithmic scale as recommended by Prophet's developers.}
    \label{tab:hyperparameters}
\end{table}

\subsection{Availability Forecast Validation} \label{availability_forecast_results}

\begin{figure}
    \centering
    \includegraphics[width=\textwidth]{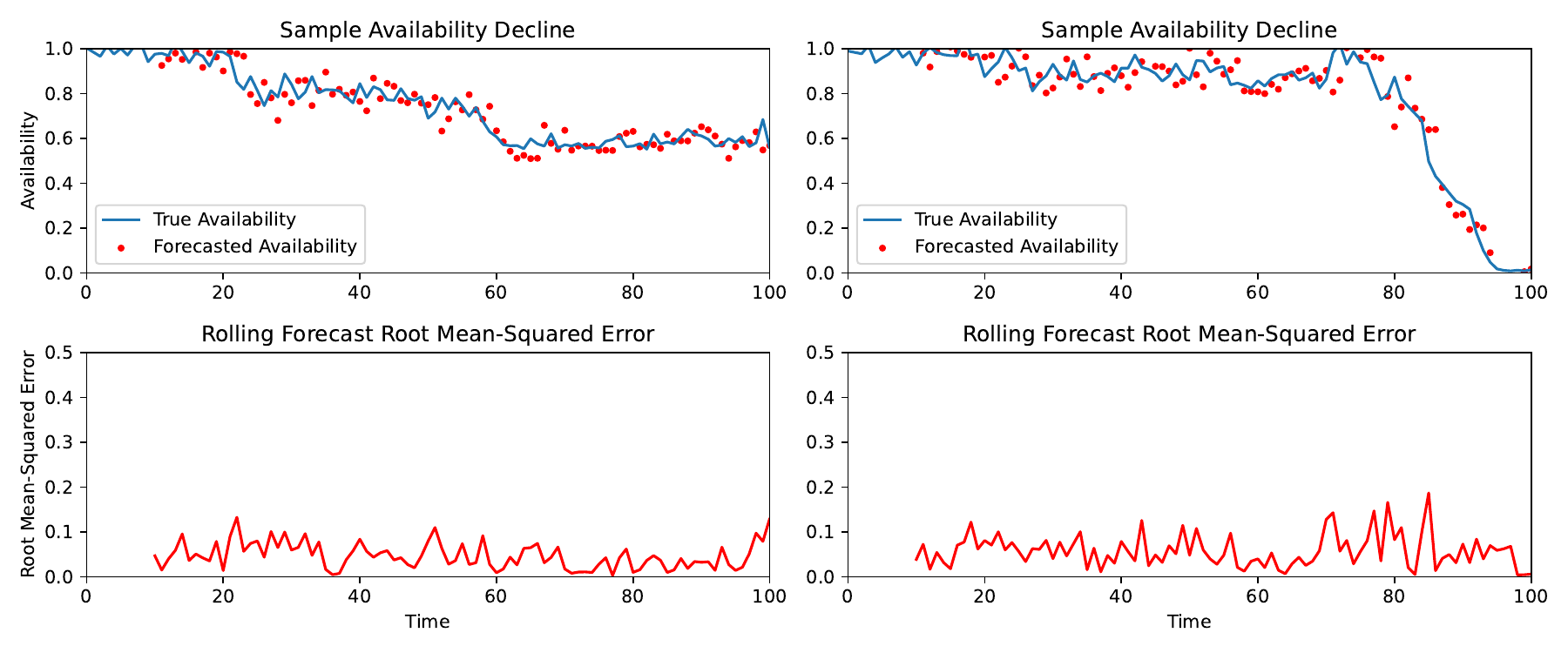}
    \caption{Availability forecasts corresponding to two situations during which a problematic vendor's availability declined. Here, we fix the rolling window size $W=10$ minutes and forecasting horizon size $R=2$ minutes. For each $M \ge 0,$ we consider a Double Exponential Smoothing model for the rolling availability time-series between the times $t_{M-10}$ and $t_{M}$  
    \textbf{Eq. \eqref{eq: Double Exponential Smoothing Equation 1}} and \textbf{Eq. \eqref{eq: Double Exponential Smoothing Equation 2}}. We then computed the rolling horizon forecast time-series between the times $t_{M+1}$ and $t_{M+2}$ minutes ahead of the window according to \textbf{Eq. \eqref{eq:Availability Forecast}}, and we computed the associated rolling horizon root mean-squared errors. The plots in the first row show the problematic vendor's actual availability (blue line) and rolling horizon predicted availability (red scatterplot) $2$ minutes ahead of each rolling window. The plots in the second row show the associated rolling horizon root mean-squared error for each rolling window.} 
    \label{fig:Sample Rolling Availability Forecasts}
\end{figure}

Recall $n_0$ is the index corresponding to a problematic vendor and that we forecast the problematic vendor's availability using a Double Exponential Smoothing model governed by \textbf{Eq. \eqref{eq: Double Exponential Smoothing Equation 1}} and \textbf{Eq. \eqref{eq: Double Exponential Smoothing Equation 2}}. We demonstrate how we validate such forecasts. Because the availability data is limited, we perform the validation on a rolling basis. We fix a positive integer $W,$ the rolling window size in minutes, and we fix a positive integer $R,$ the forecasting horizon size. For each $M \ge 0,$ we consider the historical availability rolling window time-series $\left \lbrace  a_{n,t_m} \right \rbrace_{m=M-W}^{M},$ consider a Double Exponential Smoothing model for the time-series, and compute the rolling horizon forecast time-series $\left \lbrace  \widehat{a}_{n,t_m} \right \rbrace_{m=M+1}^{M+R}$ via \textbf{Eq. \eqref{eq:Availability Forecast}}. We choose the optimal smoothing and trend factors $\alpha$ and $\eta$ in \textbf{Eqs. \eqref{eq: Double Exponential Smoothing Equation 1}} and \textbf{\eqref{eq: Double Exponential Smoothing Equation 2}}, respectively, again employing an efficient random hyperparameter search. Finally, we compute the associated rolling horizon mean-squared error between the rolling true horizon time-series $\left \lbrace  a_{n,t_m} \right \rbrace_{m=M+1}^{M+R}$ and the rolling horizon forecast time-series $\left \lbrace \widehat{a}_{n,t_m} \right \rbrace_{m=M+1}^{M+R}.$ The closer the rolling horizon mean-squared errors are to $0,$ the better the corresponding rolling horizon availability forecasts are. We show a sample true historical availability time-series, rolling horizon forecast time-series, and associated rolling horizon mean-squared errors in \textbf{Fig. \ref{fig:Sample Rolling Availability Forecasts}}. We see that the Double Exponential Smoothing models are acceptable because their associated root mean-squared errors are low.

\subsection{Wired-Off Volume Forecast Validation}

Recall when the problematic vendor $n_0$ is disabled, we use a simple linear model to represent the total wired-off volume across all enabled vendors, as presented in \textbf{Eq. \eqref{eq:Wired Off Linear Model}}. More specifically, we fit a simple linear model of the difference between the total wired-off volume and the aggregate sum of enabled vendors' expected baseline volumes against the disabled vendor's expected baseline volumes. 

\begin{figure}
    \centering
    \includegraphics[width=0.9\textwidth]{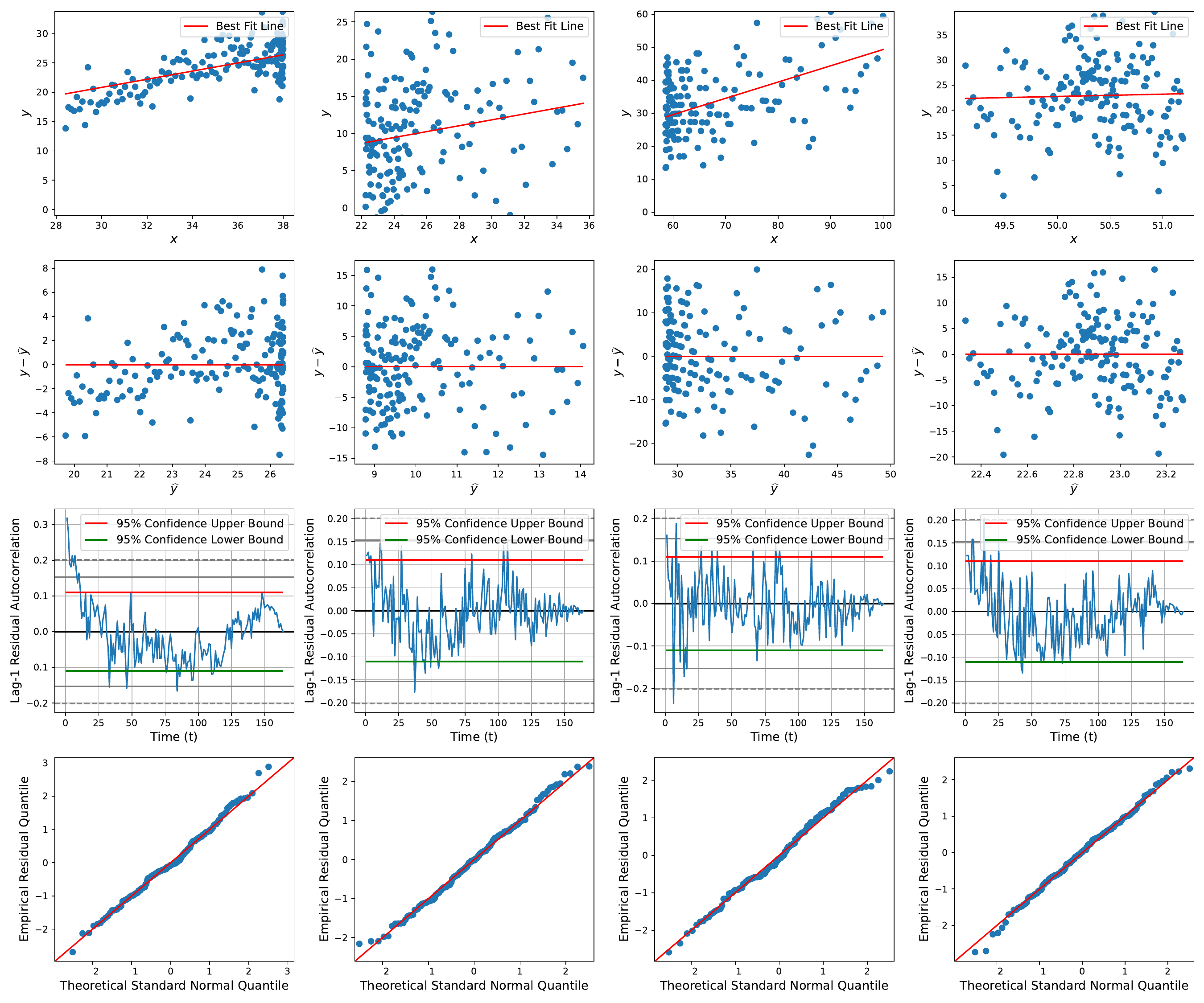}
    \caption{Three linear models corresponding to three different situations during which a problematic vendor was disabled. In the first row of plots, $x$ is the disabled vendor's expected baseline volume, while $y$ is the difference between the total volume across all enabled vendors and the total expected baseline volume across all enabled vendors. In each situation, we fit a linear model with the independent variable being $x$ and the target variable being $y.$ The second row of plots show the predicted $\widehat{y}$ value against the residual $y-\widehat{y}$.  In those plots, we see the observations are symmetrically distributed about the horizontal axis, which we would expect in a linear model. The third row of plots show the lag $1$ residual auto-correlations. In those plots, we see the majority of residual auto-correlations lie within the $95 \%$ confidence interval around $0.$ Finally, the last row of plots show the quantiles of the standard normal distribution against the quantiles of the empirical standardized residual distribution. In those plots, we see the most of the observations lie close to a straight line. Therefore, we see visual evidence pointing towards using a linear model to represent the changing volume when a vendor has been disabled. }
    \label{fig: Wired Off Vendor Experience Regression Diagnostics}
\end{figure}

We perform the regression diagnostics to validate the use of the linear model. For each $m>0,$ let $W_{n_0, t_m,\text{off}}$ and $\widehat{W}_{n_0, t_m,\text{off}}$ be the actual wired-off volume and predicted wired-off volume at time $t_m,$ respectively. For each $m>0,$ we define the residual 
    \begin{align*}
        r_{n,t_m} &= W_{n, t_m,\text{off}}- \widehat{W}_{n, t_m,\text{off}},
    \end{align*}
at time $t_m.$ We show the results of three different wired-off linear models in \textbf{Fig. \ref{fig: Wired Off Vendor Experience Regression Diagnostics}} corresponding to three different situations when a problematic vendor was actually disabled. 

Firstly, in a generic linear model, the plot of the residuals against the predicted values is symmetrically distributed about the horizontal axis. In the second row of \textbf{Fig. \ref{fig: Wired Off Vendor Experience Regression Diagnostics}}, we plot the corresponding residuals against predicted values for each wired-off linear model. We see these plots exhibit visible symmetry about the horizontal axis. 

Next, in a generic simple linear model, the \textit{recursive residuals}, which are the standardized one time-step ahead prediction errors, are independent and identically Gaussian distributed. In order to validate this for our sample wired-off linear models, we formally perform the Harvey-Collier test \citep{HarveyCollier1977} at a fixed percent significance level. In this test, the null hypothesis states the recursive residuals are normally distributed. Therefore, we need to check whether we do not reject the null hypothesis. We compute Harvey-Collier test statistics corresponding to our sample wired-off linear models and tabulate their $p$-values in the first column of \textbf{Table \ref{tab: Linear Model HC and DW Statistics}}. At the $5$ percent significance level, since the $p$-values are much greater than $5$ percent, we do not reject the Harvey-Collier null hypothesis in any of our wired-off linear models.

Next, in a generic simple linear model, the majority of auto-correlations at lag $1$ lies within the $95$ percent confidence bound about $0$ i.e. approximately within $\left [ -\sqrt{2/M}, \sqrt{2/M}\right],$ where $M$ is the number of observations. Moreover, we formally perform the Durbin-Watson test \citep{DurbinWatson1971}, whose null hypothesis states there is no auto-correlation within the residuals. In a generic linear model, the residuals are independent, and so we need to check whether we do not reject the null hypothesis. A heuristic condition to not reject the null hypothesis is if the Durbin-Watson test statistic lies in the interval $[1.5,3.5].$ We tabulate the Durbin-Watson test statistics corresponding to our sampled wired-off linear models in the second column of \textbf{Table \ref{tab: Linear Model HC and DW Statistics}}. Because all our three obtained test statistics lie between $1.5$ and $3.5,$  we do not reject the Durbin-Watson null hypothesis in any of our sample wired-off linear models.

Lastly, in a generic linear model, the residuals are normally distributed. To validate this, we inspect the quantiles of the standardized residuals against the theoretical quantiles of the standard normal distribution and check whether the observations lie close to the straight line $y=x.$  For our sample wired-off linear models, we plot the corresponding quantiles of the standardized residuals against the theoretical quantiles of the standard normal distribution in the last row of \textbf{Fig. \ref{fig: Wired Off Vendor Experience Regression Diagnostics}}. We see these plots exhibit visible symmetry.

\begin{table}
\begin{center}
\begin{tabular}{|c c|} 
 \hline
 $p_{\text{HC}}$ & $t_{\text{DW}}$ \\ [0.5ex] 
 \hline\hline
 $0.51$ & $1.72$ \\
 \hline
 $0.49$ & $1.68$ \\
 \hline
 $0.65$ & $1.73$ \\
 \hline
\end{tabular}
\caption{Harvey-Collier and Durbin-Watson test results obtained after fitting three linear models corresponding to the three different situations during which a problematic vendor was disabled, as described in \textbf{Fig. \ref{fig: Wired Off Vendor Experience Regression Diagnostics}}. Each row has the results of a particular linear model. The first column contains the three $p$-values corresponding to the performed Harvey-Collier tests. Since all $p$-values are much greater than $5$ percent, we do not reject the Harvey-Collier null hypothesis at the $5$ percent significance level for each linear model. The second column contains the three test statistics corresponding to the Durbin-Watson tests performed. Since all test statistics lie between $1.5$ and $3.5,$ we do not reject the Durbin-Watson null hypothesis for each linear model. Therefore, we have strong evidence validating the use of a linear model to represent the wired-off customer behavior when a problematic vendor has been disabled.}
\end{center}
\label{tab: Linear Model HC and DW Statistics}
\end{table}

\subsection{Wire-off Recommendation}

With the forecasts $\left \lbrace \widehat{W}_{n,t_m, \text{off}} \right \rbrace_{m = 1}^R
$  and $\left \lbrace \widehat{W}_{n,t_m, \text{on}} \right \rbrace_{m = 1}^R, $ where $R$ is the fixed future horizon size defined in \textbf{Section \ref{availability_forecast_results}}, we determine the smallest index $m^*$ in which  

\begin{equation}
    \widehat{W}_{n,t_{m}, \text{off}} > \widehat{W}_{n,t_{m}, \text{on}}
    \label{eq:optimal_wireoff}
\end{equation}
for all $m \ge m^*$ in our simulated window. If $m^*$ exists, we make a decision to wire off vendor $n$ starting at time $t_{m^*}$. Otherwise, we keep vendor $n$ wired on.

\begin{figure}
    \centering
    \includegraphics[width=0.8\textwidth]{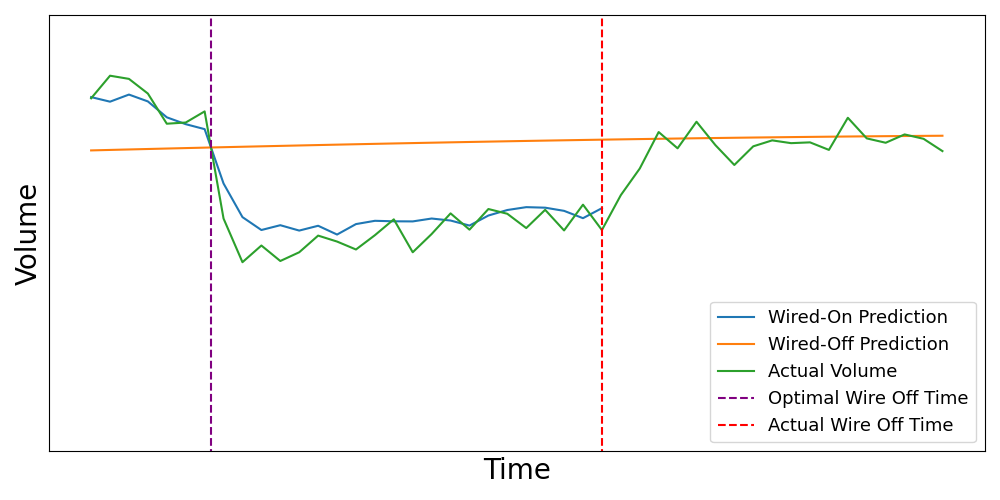}
    \caption{The simulation results for a real vendor incident in which the problematic vendor was wired off. The wired-on simulation is only run until the actual wire off time; real-time availability trends are used for the wired-on simulation, which are not available while the vendor is wired off. Both the wired-on and wired-off predictions accurately recreate the actual completed experiences observed under the respective conditions. The purple dotted line shows the optimal wire off time according to \textbf{Eq. \eqref{eq:optimal_wireoff}}; between this time and our actual wire off time, the actual volume is less than our wired-off prediction, indicating that fewer customers were able to complete their experiences than otherwise would have been possible if the vendor were wired off.}
    \label{fig:simulation-result}
\end{figure}

\textbf{Figure \ref{fig:simulation-result}} shows our simulated trends for a real incident in which a vendor was experiencing issues. The optimal wire off time according to \textbf{Eq. \eqref{eq:optimal_wireoff}} is considerably earlier than the actual wire off time. We define $t_{\text{actual}}$ to be the actual wire-off time and $t_{m^*}$ to be the optimal wire off time based on \textbf{Eq. \eqref{eq:optimal_wireoff}} and our simulation. \textbf{Table \ref{tab:lead_times}} shows the lead time $t_{\text{actual}} - t_{m^*}$ of our optimal simulated wire off time over the actual wire off time for several historical incidents. We see that our optimal wire off time is often several minutes before the actual wire off time. Depending on the experience and time of day, this can equate to a considerable number of additional completed customer experiences.

\begin{table}
    \centering
    \begin{tabular}{|c | c|}
        \hline
        Incident & $t_{\text{actual}} - t_{m^*}$ [min] \\
        \hline \hline
        1 & 11 \\
        \hline
        2 & 16 \\ 
        \hline
        3 & 10 \\
        \hline
        4 & 12 \\
        \hline
        5 & 3 \\ 
        \hline
        6 & 11 \\
        \hline
        7 & 10 \\
        \hline
    \end{tabular}
    \caption{For seven different incidents during which a particular vendor experienced an outage, we tabulate the difference in the optimal wire off time $t_{m^*}$ according to our methodology and the actual wire off time $t_{\text{actual}}$. In all situations, we observe the optimal wire off time is earlier than the actual wire off time.}
    \label{tab:lead_times}
\end{table}

Thus, using our approach, we can see benefits to both our customers and the business over the existing solution. In the short term, we are immediately providing an improved experience which reduces friction for our customers. In the long term, the cumulative improved experiences will grow over several vendor incidents; this not only means more overall customer experiences completed, but also greater customer satisfaction over time, which will drive people to continue using eBay.

%%%%%%%%%%%%%%%%%%%%%%%%%%%%%%%%%%%%%%%%%%%%%%%

\section{Future Work}
In this paper, we predicted customer behavior to arrive at the decision to wire off a problematic vendor. A natural question to ask is when a disabled vendor should be enabled again for customers. We hope to use the methodology in this paper to arrive at a data-driven answer to this question. This problem poses several additional challenges; namely, we would only like to re-enable a vendor when we know its impact has been mitigated. However, after the vendor is wired off, we no longer see real-time metrics and must therefore find another way to measure its availability. A solution for this may depend on the data available for particular customer experience and/or vendor. 

Furthermore, recall the key assumption for our approach to forecasting completed customer experiences is the existence of only one problematic vendor. In many situations, however, more than one vendor could experience issues, which would severely impact eBay's business. We hope to modify the probabilistic model in \textbf{Section \ref{probabilistic_model}} in order to forecast customer behavior when multiple problematic vendors are either wired on or wired off.

%%%%%%%%%%%%%%%%%%%%%%%%%%%%%%%%%%%%%%%%%%%%%%%

% \bibliographystyle{unsrt} 
\bibliographystyle{elsarticle-harv}
\bibliography{main}

\begin{thebibliography}{9}
\expandafter\ifx\csname natexlab\endcsname\relax\def\natexlab#1{#1}\fi
\providecommand{\url}[1]{\texttt{#1}}
\providecommand{\href}[2]{#2}
\providecommand{\path}[1]{#1}
\providecommand{\DOIprefix}{doi:}
\providecommand{\ArXivprefix}{arXiv:}
\providecommand{\URLprefix}{URL: }
\providecommand{\Pubmedprefix}{pmid:}
\providecommand{\doi}[1]{\href{http://dx.doi.org/#1}{\path{#1}}}
\providecommand{\Pubmed}[1]{\href{pmid:#1}{\path{#1}}}
\providecommand{\bibinfo}[2]{#2}
\ifx\xfnm\relax \def\xfnm[#1]{\unskip,\space#1}\fi
%Type = Article
\bibitem[{Bergstra and Bengio(2012)}]{BergstraBengio}
\bibinfo{author}{Bergstra, J.}, \bibinfo{author}{Bengio, Y.},
  \bibinfo{year}{2012}.
\newblock \bibinfo{title}{Random search for hyper-parameter optimization}.
\newblock \bibinfo{journal}{Journal of Machine Learning Research}
  \bibinfo{volume}{13}, \bibinfo{pages}{281--305}.
\newblock \URLprefix \url{http://jmlr.org/papers/v13/bergstra12a.html}.
%Type = Inproceedings
\bibitem[{Brown(1957)}]{brown1957exponential}
\bibinfo{author}{Brown, R.G.}, \bibinfo{year}{1957}.
\newblock \bibinfo{title}{Exponential smoothing for predicting demand}, in:
  \bibinfo{booktitle}{Operations Research}, \bibinfo{organization}{Inst
  Operations Research Management Sciences}. pp. \bibinfo{pages}{145--145}.
%Type = Article
\bibitem[{Durbin and Watson(1971)}]{DurbinWatson1971}
\bibinfo{author}{Durbin, J.}, \bibinfo{author}{Watson, G.S.},
  \bibinfo{year}{1971}.
\newblock \bibinfo{title}{Testing for serial correlation in least squares
  regression. iii}.
\newblock \bibinfo{journal}{Biometrika} \bibinfo{volume}{58},
  \bibinfo{pages}{1--19}.
\newblock \URLprefix \url{http://www.jstor.org/stable/2334313}.
%Type = Book
\bibitem[{Greene(2003)}]{Greene_2003}
\bibinfo{author}{Greene, W.H.}, \bibinfo{year}{2003}.
\newblock \bibinfo{title}{Econometric analysis}.
\newblock \bibinfo{edition}{5} ed., \bibinfo{publisher}{Prentice Hall}.
%Type = Article
\bibitem[{Harvey and Collier(1977)}]{HarveyCollier1977}
\bibinfo{author}{Harvey, A.}, \bibinfo{author}{Collier, P.},
  \bibinfo{year}{1977}.
\newblock \bibinfo{title}{Testing for functional misspecification in regression
  analysis}.
\newblock \bibinfo{journal}{Journal of Econometrics} \bibinfo{volume}{6},
  \bibinfo{pages}{103--119}.
\newblock \URLprefix
  \url{https://EconPapers.repec.org/RePEc:eee:econom:v:6:y:1977:i:1:p:103-119}.
%Type = Article
\bibitem[{Holt(2004)}]{holt2004forecasting}
\bibinfo{author}{Holt, C.C.}, \bibinfo{year}{2004}.
\newblock \bibinfo{title}{Forecasting seasonals and trends by exponentially
  weighted moving averages}.
\newblock \bibinfo{journal}{International Journal of Forecasting}
  \bibinfo{volume}{20}, \bibinfo{pages}{5--10}.
\newblock \URLprefix
  \url{https://www.sciencedirect.com/science/article/pii/S0169207003001134},
  \DOIprefix\doi{https://doi.org/10.1016/j.ijforecast.2003.09.015}.
%Type = Article
\bibitem[{Koehler et~al.(2001)Koehler, Snyder and
  Ord}]{multiplicativeSeasonality}
\bibinfo{author}{Koehler, A.B.}, \bibinfo{author}{Snyder, R.D.},
  \bibinfo{author}{Ord, J.}, \bibinfo{year}{2001}.
\newblock \bibinfo{title}{Forecasting models and prediction intervals for the
  multiplicative holt–winters method}.
\newblock \bibinfo{journal}{International Journal of Forecasting}
  \bibinfo{volume}{17}, \bibinfo{pages}{269--286}.
\newblock \URLprefix
  \url{https://www.sciencedirect.com/science/article/pii/S0169207001000814},
  \DOIprefix\doi{https://doi.org/10.1016/S0169-2070(01)00081-4}.
%Type = Article
\bibitem[{{National Institute of Standards and
  Technology}(2012)}]{NistDoubleExponential}
\bibinfo{author}{{National Institute of Standards and Technology}},
  \bibinfo{year}{2012}.
\newblock \bibinfo{title}{{6.4.3.3. Double Exponential Smoothing}}.
\newblock \bibinfo{journal}{NIST/SEMATECH e-Handbook of Statistical Methods}
  \URLprefix
  \url{https://www.itl.nist.gov/div898/handbook/pmc/section4/pmc433.htm},
  \DOIprefix\doi{https://doi.org/10.18434/M32189}.
%Type = Article
\bibitem[{Taylor and Letham(2018)}]{TaylorLetham2018}
\bibinfo{author}{Taylor, S.J.}, \bibinfo{author}{Letham, B.},
  \bibinfo{year}{2018}.
\newblock \bibinfo{title}{Forecasting at scale}.
\newblock \bibinfo{journal}{The American Statistician} \bibinfo{volume}{72},
  \bibinfo{pages}{37--45}.
\newblock \URLprefix \url{https://doi.org/10.1080/00031305.2017.1380080},
  \DOIprefix\doi{10.1080/00031305.2017.1380080},
  \href{http://arxiv.org/abs/https://doi.org/10.1080/00031305.2017.1380080}{{\tt
  arXiv:https://doi.org/10.1080/00031305.2017.1380080}}.

\end{thebibliography}

\end{document}